\documentclass[10pt,twocolumn,letterpaper]{article}

\usepackage[table,dvipsnames]{xcolor}
\definecolor{lightpink}{RGB}{255, 230, 230}
\definecolor{lightblue}{RGB}{230, 240, 255}

\usepackage{multirow}
\usepackage{cvpr}      %

\definecolor{cvprblue}{rgb}{0.21,0.49,0.74}
\usepackage[pagebackref,breaklinks,colorlinks,allcolors=cvprblue]{hyperref}

\title{
\vspace{-3mm}
Stag-1: Towards Realistic 4D Driving Simulation with Video Generation Model}

\author{First Author\\
Institution1\\
Institution1 address\\
{\tt\small firstauthor@i1.org}
\and
Second Author\\
Institution2\\
First line of institution2 address\\
{\tt\small secondauthor@i2.org}
}

\begin{document}

\newcommand*\samethanks[1][\value{footnote}]{\footnotemark[#1]}
\DeclareUrlCommand\url{\color{magenta}}

\author{\textbf{Lening Wang}$^{1,2,4,}$\footnotemark[1] $^,$\footnotemark[2] \quad 
\textbf{Wenzhao Zheng}$^{2,}$\footnotemark[1] $^,$\footnotemark[3]  \quad
\textbf{Dalong Du}$^{2,3}$\quad
\textbf{Yunpeng Zhang}$^3$  \quad  
\textbf{Yilong Ren}$^1$ \\ \quad
\textbf{Han Jiang}$^{1}$  \quad
\textbf{Zhiyong Cui}$^{1}$\quad 
\textbf{Haiyang Yu}$^{1}$\quad 
\textbf{Jie Zhou}$^{2}$ \quad 
\textbf{Jiwen Lu}$^{2}$ \quad 
\textbf{Shanghang Zhang}$^{4}$ \\
Project Page: \url{https://wzzheng.net/Stag}\\
Large Driving Models: \url{https:/github.com/wzzheng/LDM}\\
$^1$Beihang University, \quad\quad 
$^2$Tsinghua University \quad\quad
$^3$PhiGent Robotics  \quad\quad  
$^4$ Peking University \\
\texttt{leningwang@buaa.edu.cn; wenzhao.zheng@outlook.com}
\vspace{-5mm}
}

\twocolumn[{%
\renewcommand\twocolumn[1][]{#1}%

\maketitle
\begin{center}
    \centering
    \includegraphics[width=1\linewidth]{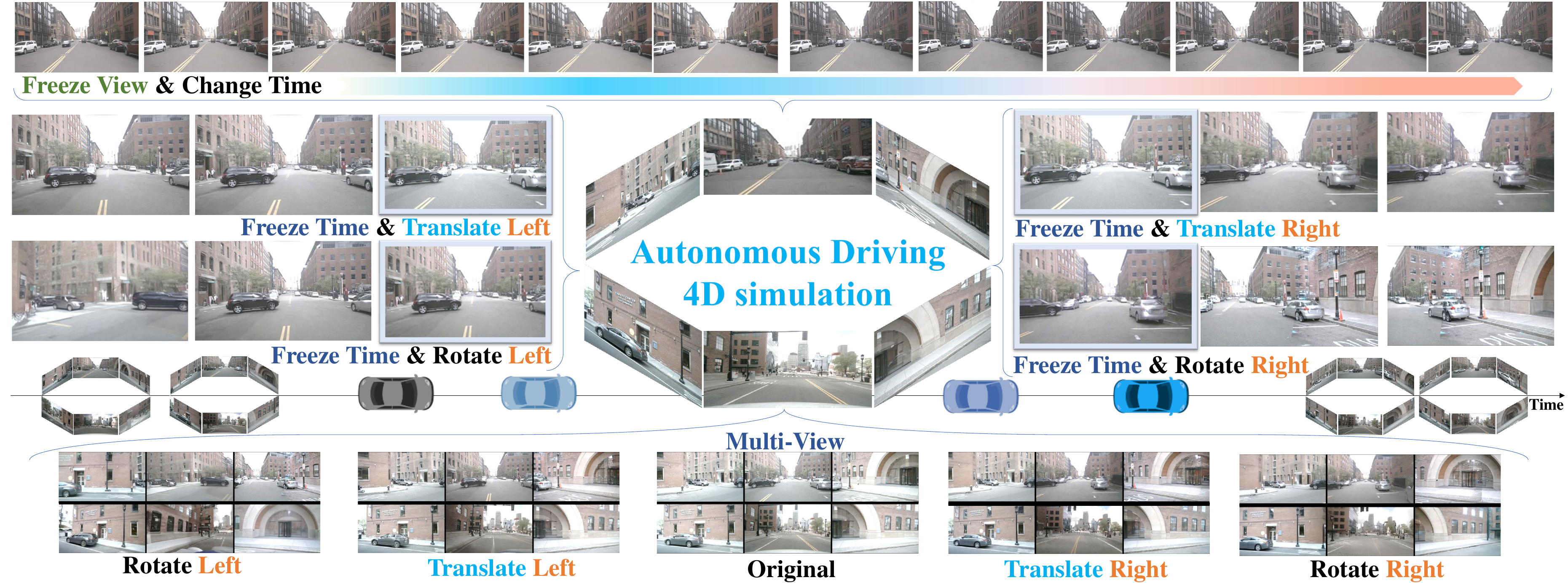}
    
    \captionof{figure}{\textbf{Spatial-Temporal simulAtion for drivinG (Stag-1)} enables controllable 4D autonomous driving simulation with spatial-temporal decoupling. Stag-1 can decompose the original spatial-temporal relationships of real-world scenes to enable controllable autonomous driving simulation. This allows for adjustments such as fixing the camera viewpoint while advancing time or translating and rotating space while keeping time stationary. Additionally, Stag-1 maintains synchronized variations across six panoramic views.
}
\label{pic-main00}
\end{center}%
}]

\renewcommand{\thefootnote}{\fnsymbol{footnote}}
\footnotetext[1]{Equal contribution. $\dagger$Work done during an internship at Tsinghua University and Peking University. $\ddagger$Project leader. 
}
\renewcommand{\thefootnote}{\arabic{footnote}}

\begin{abstract}

4D driving simulation is essential for developing realistic autonomous driving simulators. Despite advancements in existing methods for generating driving scenes, significant challenges remain in view transformation and spatial-temporal dynamic modeling. To address these limitations, we propose a \textbf{Spatial-Temporal simulAtion for drivinG (Stag-1)} model to reconstruct real-world scenes and design a controllable generative network to achieve 4D simulation. Stag-1 constructs continuous 4D point cloud scenes using surround-view data from autonomous vehicles. It decouples spatial-temporal relationships and produces coherent keyframe videos. Additionally, Stag-1 leverages video generation models to obtain photo-realistic and controllable 4D driving simulation videos from any perspective. To expand the range of view generation, we train vehicle motion videos based on decomposed camera poses, enhancing modeling capabilities for distant scenes. Furthermore, we reconstruct vehicle camera trajectories to integrate 3D points across consecutive views, enabling comprehensive scene understanding along the temporal dimension. Following extensive multi-level scene training, Stag-1 can simulate from any desired viewpoint and achieve a deep understanding of scene evolution under static spatial-temporal conditions. Compared to existing methods, our approach shows promising performance in multi-view scene consistency, background coherence, and accuracy, and contributes to the ongoing advancements in realistic autonomous driving simulation.
Code: \url{https://github.com/wzzheng/Stag}.

\end{abstract}    
\vspace{-4mm}
\section{Introduction}
\label{sec:intro}
\vspace{-1mm}

As autonomous driving capabilities advance in perception \cite{liu2023petrv2, li2022bevformer,liu2024fmdnet,liu2024glmdrivenet}, prediction \cite{gao2020vectornet, zhang2022beverse}, and planning \cite{jiang2023vad, zheng2025occworld}, significant progress has also been made in end-to-end networks \cite{hu2023planning, hu2022stp3}. With these advancements, comprehensive testing and validation of autonomous vehicles are increasingly critical \cite{feng2023dense}. However, real-world vehicle testing remains time-consuming, costly, and limited in scenario coverage.

Mainstream research increasingly relies on simulation software for extensive algorithm testing and validation \cite{shao2024lmdrive}. Yet, simulations based on 3D modeling struggles to accurately replicate realistic driving scenarios, creating a substantial gap between synthetic environments and real-world conditions \cite{zhou2024simgen}.
To address this, existing autonomous driving testing solutions strive to build highly realistic scenarios for validating driving algorithms \cite{wang2023drivedreamer, gao2024vista}. With the rapid advancement of text-to-image and text-to-video generative models \cite{zhang2023adding}, some research has focused on generating trajectory-controlled images or videos to simulate autonomous driving scenes, guided by maps and surrounding vehicle poses to improve scene accuracy \cite{kim2021drivegan,gao2023magicdrive}. However, real-world driving involves constantly moving pedestrians, vehicles, and objects that introduce structural changes to the environment. Video generation methods often struggle to capture these dynamic changes or the close interactions between elements, leading to inconsistencies in scene continuity, such as background and vehicle type shifts, which complicate maintaining temporal consistency \cite{zhao2024drivedreamer4d,hu2023gaia}. Recently, approaches based on NeRF \cite{wei2024editable} and 3D Gaussian Splatting (3DGS) \cite{huang2024textit,gao2024magicdrive3d} have aimed to capture dynamic elements with greater precision by rendering and modeling 3D scenes. Nonetheless, they still face challenges in reconstructing 4D scenes from arbitrary viewpoints, handling extensive dynamic view changes with significant camera movement, and managing long-term temporal transformations under static views.

\begin{figure*}[t]
  \vspace{-5mm}
  \centering
  \includegraphics[width=1\linewidth]{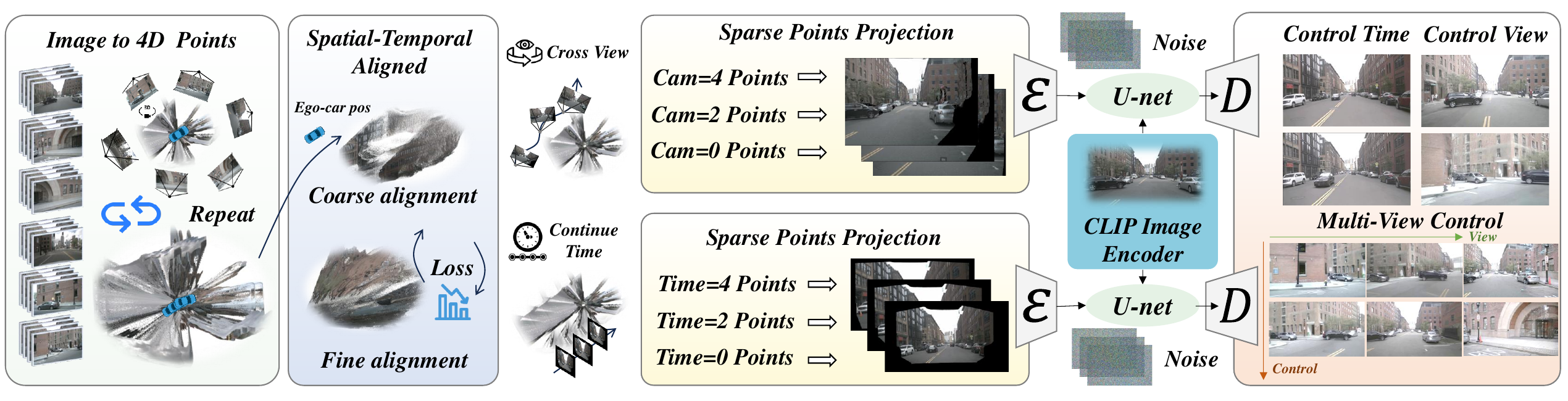}
  \vspace{-9mm}
  \caption{\textbf{Our Stag-1 framework is a 4D generative model for autonomous driving simulation.} It reconstructs 4D scenes from point clouds and projects them into continuous, sparse keyframes. A spatial-temporal fusion framework is then used to generate simulation scenarios. Two key design aspects guide our approach: 1) We develop a method for 4D point cloud matching and keyframe reconstruction, ensuring the accurate generation of continuous, sparse keyframes that account for both vehicle motion and the need for spatial-temporal decoupling in simulation. 2) We build a spatial-temporal fusion framework that integrates surround-view information and continuous scene projection to ensure accurate simulation generation.}
  \label{pic-main01-1}
\vspace{-5mm}
\end{figure*}

To enable more realistic autonomous driving testing, we propose \textbf{Spatial-Temporal simulAtion for drivinG (Stag-1)}, a controllable 4D simulation framework based on real-world autonomous driving scenes, as shown in Figure \ref{pic-main01-1}. Our approach begins by constructing 3D point clouds frame-by-frame using surround-view data from autonomous vehicles. We then develop a rough alignment network based on ego-car and camera parameters. Next, we iteratively refine the point clouds and align them with sequential scenes, resulting in a 4D point cloud that incorporates both camera and vehicle motion parameters. This process accurately captures the 4D structure of real-world environments. Furthermore, we develop a multi-view interaction-based sparse point cloud completion network, which allows for controllable 4D simulation video synthesis in autonomous driving applications. To improve the quality of continuous scene simulation, we also design a cross-view diffusion-based generative network that addresses two key challenges: comprehensive dynamic viewpoint modeling in static scenes and precise static viewpoint modeling in dynamic scenes. This network compensates for missing information in sparse point clouds, ensuring continuity in scene transitions and accuracy in temporal changes.

\section{Related Work}
\label{sec:Related Work}

\textbf{Scene Simulation for Autonomous Driving.} In early autonomous driving simulation tasks, creating realistic street views often required manual 3D scene modeling, which resulted in significant gaps from real-world scenarios \cite{dosovitskiy2017carla, wang2021simulation, wei2024editable}. The advent of NeRF \cite{mildenhall2021nerf} introduced a novel approach by reconstructing real scenes from multi-view images, opening new possibilities for autonomous driving simulation \cite{lindstrom2024nerfs, chen2024s, tonderski2024neurad}. Later, the introduction of 3DGS \cite{kerbl20233d} methods further enhanced both the effectiveness and efficiency of 3D scene reconstruction. Some researchers \cite{zhou2024simgen, gao2024magicdrive3d, huang2024textit, zhao2024drivedreamer4d} have explored combining autonomous driving image scene generation with 3DGS to improve scene coherence. However, methods based on NeRF and 3DGS struggle with large camera viewpoint shifts and lack precise control over temporal progression from fixed viewpoints, limiting their application in controlled autonomous driving simulations. In contrast, Stag-1 combines realistic sparse point cloud representation in 4D reconstruction with video generation using diffusion models, achieving both cross-view and temporal consistency. This enables controllable scene generation from any viewpoint at any time, advancing 4D simulation tasks for autonomous driving.

\textbf{Scene Generation for Autonomous Driving.} Generative adversarial networks \cite{goodfellow2014generative} have made significant advances in image generation, and with the emergence of diffusion architectures \cite{rombach2022high, peebles2023scalable, croitoru2023diffusion, xing2025dynamicrafter, blattmann2023stable}, numerous studies have demonstrated powerful capabilities in image and video synthesis. To enhance the controllability of generation, subsequent networks have been refined to enable condition-based video generation using inputs such as text and images \cite{zhao2024uni, li2025controlnet, ceylan2023pix2video}. In the field of autonomous driving, scene generation \cite{kim2021drivegan, lu2025wovogen} plays a crucial role in enhancing the system's adaptability to diverse driving scenarios and facilitating closed-loop testing \cite{shao2024lmdrive, caesar2021nuplan, yang2024drivearena}. As such, trajectory control \cite{wang2023drivedreamer, gao2024vista, hu2023gaia} or the use of 3D bounding boxes and scene description text \cite{gao2023magicdrive, gao2024magicdrive3d} have been introduced to guide autonomous driving scene generation. However, despite the inclusion of various control signals, existing methods still face significant challenges in maintaining scene consistency and ensuring high controllability. In contrast, Stag-1 demonstrates clear advantages in maintaining consistency across continuous scenes, providing arbitrary viewpoint and time control, and ensuring multi-view consistency.

\textbf{Controllable Video Generation.} With advancements in text-to-image \cite{li2019controllable, ramesh2021zero} and image-to-video generation models \cite{xing2025dynamicrafter, blattmann2023stable, karras2023dreampose}, enhancing the controllability of generative models has garnered significant attention. Diverse condition-control \cite{li2025controlnet} networks now enable controllable image or video generation using inputs such as sketches \cite{xing2023diffsketcher}, trajectories \cite{yu2024viewcrafter, giannone2023aligning}, and more. In research focused on generative modeling for autonomous driving, much emphasis has been placed on vehicle trajectory control \cite{gao2024vista, hu2023gaia, wang2023drivedreamer} or on using target bounding boxes and maps for guidance \cite{gao2023magicdrive, kim2021drivegan}. However, these approaches often lack a comprehensive understanding of 4D spatial relationships, making it challenging to accurately capture and distinguish spatial-temporal dynamics. In contrast, our work establishes continuous 4D point clouds to decouple spatial and temporal relationships in generative tasks, enhancing editing capabilities for generative simulations in autonomous driving.

\section{Proposed Approch}

\begin{figure*}[t]
  \centering
  \includegraphics[width=1\linewidth]{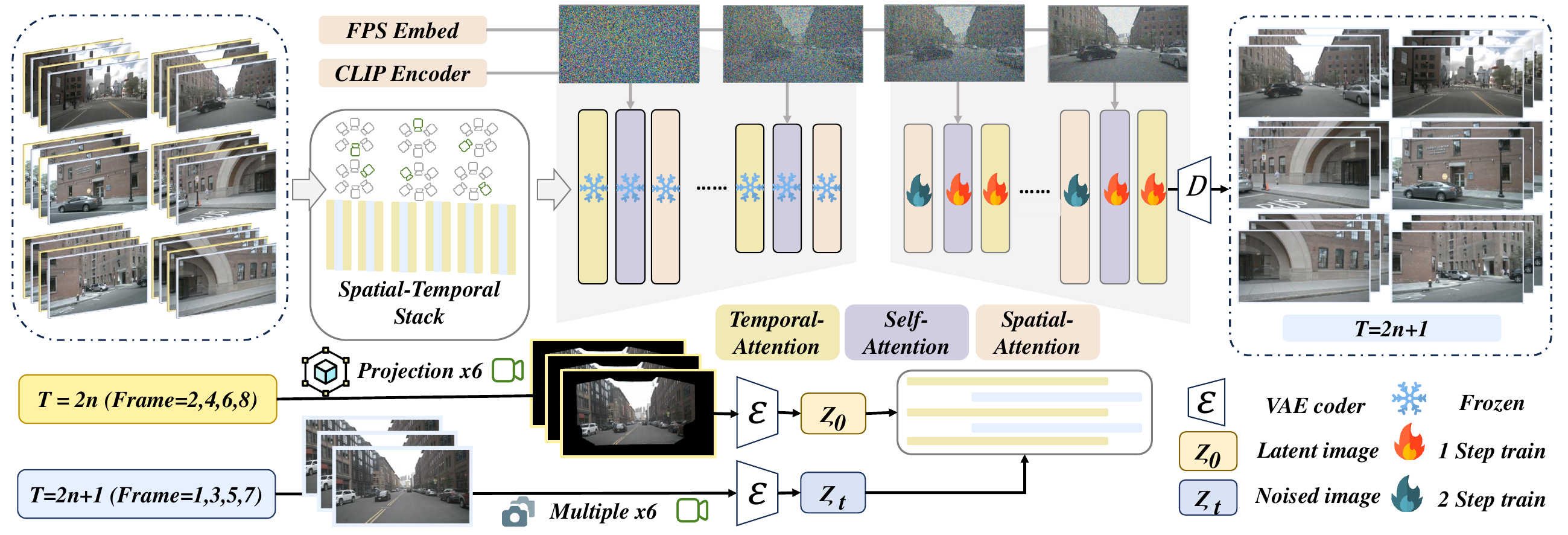}
    \vspace{-8mm}
  \caption{\textbf{The Stag-1 training framework pipeline is designed in two Stages.} In the time-focused Stage, we use even keyframes from a single viewpoint to generate a 4D point cloud, which is then projected with odd keyframe parameters as conditions, with the odd keyframes serving as labels for training. In the spatial-focused Stage, surround-view information is incorporated to extract inter-image features from the surrounding viewpoints, followed by the training of the spatial-temporal block.}
  \label{pic-main02-1}
\vspace{-5mm}
\end{figure*}

\subsection{4D Autonomous Driving Simulation}
\label{sec:4D Autonomous Driving Simulation}

Generative 4D autonomous driving simulation aims to address the lack of realism in traditional autonomous driving simulation scenarios \cite{dosovitskiy2017carla} and overcome the limitations of image generation models in terms of scene quality \cite{gao2023magicdrive} and control capabilities \cite{gao2024vista}. Formally, the generative 4D autonomous driving simulation generates a scene $G_{sim}\in \mathbb{R} ^{H\times W\times t\times 3}$ that based on the real-world scene $S_{gt}\in \mathbb{R} ^{H\times W\times t\times 3}$ and a set of control signals $C_s$, can be expressed as: 
\vspace{-2mm}
\begin{equation}
G_{sim}(C_s, S_{gt}) = f(S_{gt}, C_s), 
\vspace{-2mm}
\end{equation} where $C_s = \{C_{p}, T_t, \dots, V_s\}$ is the set of control signals (such as camera poses $C_{p}$, time $T_t$, and vehicle states $V_s$).

The traditional approach typically includes control signals $C$ such as the vehicle's continuous trajectory $T$, throttle position $\theta$, steering angle $\delta$, and the real positions of surrounding vehicles and BEV maps $M_{bev}$, which control the scene $G_{sim}$. This can be expressed as:
\vspace{-2mm}
\begin{equation}
C_g= \{ T, \theta, \delta, P_s, M_{bev} \}.
\vspace{-2mm}
\end{equation}

However, the traditional methods often fail to effectively capture the true relationship between time and space in the scene, leading to temporal jumps and a lack of controllability in the generated scenes. Therefore, we explore 4D scene point cloud reconstruction for realistic scene understanding and high-quality image generation using generative models. The method combines 4D point cloud, camera parameters, and temporal information, and uses a generation framework to effectively capture the independent variations of time and space, enabling more natural and precise autonomous driving simulations, which can be expressed as: 
\vspace{-2mm}
\begin{equation}
G_{sim}(C_{p}, T_t) = f(C_{p}, T_t, \{S_{gt}, V_s, C_{\theta}\}), 
\vspace{-2mm}
\end{equation} where $C_{\theta}$ represents the original parameters of the ego vehicle and cameras. In this way, we can generate realistic simulated scenes consistent with the control signals.

To accurately control the scene, we extract the 4D point cloud $P_t^{a}$ from the current scene $S_{gt}$, and project $P_t^{a}$ onto continuous 2D images $P_i^{2D}(t_k)$ under the continuous conditions of $C_{p}$ and $T_t$, which forms a keyframe video $V_{key} \in \mathbb{R}^{H \times W \times t \times 3}$. Then, we use a video generation network $G\left( \cdot \right)$ to generate a continuous, accurate, and controllable 4D autonomous driving simulation scene, $G_{sim}$.

The process includes two key steps: reconstructing an accurate 4D point cloud $P_t^{a}$ and projecting the 4D point cloud into keyframes $V_{key}$ based on control information for generation. We will provide a detailed explanation of these two steps in Section \ref{sec:4D_scene_reconstruction} and Section \ref{sec:4D simulation}.

\subsection{Spatial-Temporal Aligned Reconstruction}%
\label{sec:4D_scene_reconstruction}

The construction of generative 4D autonomous driving simulation scenes $G_{sim}$ relies on an accurate 4D point cloud \( P_t^{a} \). Based on the practical needs of autonomous driving, we define the form of \( P_t^{a} \) according to three principles: 1) Authenticity: The 4D point cloud must be constructed with real parameters, requiring accurate scene size and range assessment, rather than just relative proportions. 2) Accuracy: The scene should precisely estimate object positions and distances to enhance 3D point cloud precision. 3) Consistency: Each scene frame should align with the vehicle or camera parameters for coherence.

Following these principles, we first estimate and use surround-view camera parameters $R^{(t, v)}$, $T^{(t, v)}$ to generate a surround-view 3D point cloud $P_{3D}$. We further align the point cloud using ego vehicle parameters $R_t$, $T_t$, refining it iteratively to build an accurate 4D point cloud scene \( P_t^{a} \).

\textbf{Single-Frame 3D Point Cloud Construction.} Specifically, for constructing a 3D scene from a single frame, we process each image \( S_{gt}^{(t, v)} \in \mathbb{R}^{H \times W \times 3 \times t \times V} \), where \( V = 6 \) represents six surround-views. Following part of the R3D3 \cite{schmied2023r3d3} approach for per-frame depth estimation \( D_{im}^{(t, v)} \in \mathbb{R}^{H \times W \times t \times V} \).

Then, we use the corresponding camera pose \( R^{(t, v)} \) and \( T^{(t, v)} \) to get an accurate surround-view point cloud \( P^{(t, v)} \). By combining the point clouds from all views \( v \), we obtain the surround-view point cloud \( P_t \) at time \( t \), which can be expressed as:
\vspace{-2mm}
\begin{equation}
 P_t = \bigcup_{v=1}^V P^{(t, v)} = \bigcup_{v=1}^V (R^{(t, v)} \cdot D_{im}^{(t, v)} + T^{(t, v)}).
 \vspace{-2mm}
 \end{equation} 

\textbf{Continue-Frame 4D Point Cloud Coarse Alignment.} For each \( P_t \in \mathbb{R}^{N_t \times 3} \), where \( \mathbf{p}_i = [x_i, y_i, z_i]^T \) represents the coordinates of the \( i \)-th point in the local coordinate system of the ego vehicle, we also need to perform continuous-frame 4D point cloud alignment. We apply the following transformation:
\vspace{-2mm}
\begin{equation}
 \mathbf{p}_i^{a}(t) = R_t \cdot \mathbf{p}_i + T_t, 
 \vspace{-2mm}
 \end{equation} where \( \mathbf{p}_i^{a}(t) \) represents the transformed 3D point in the global coordinate system, \( R_t \in \mathbb{R}^{3 \times 3} \) and \( T_t \in \mathbb{R}^3 \)  represents vehicle's pose at time \( t \).

Then, to construct a complete 4D point cloud, we aligned 4D point cloud sequence at time step \( t \), denoted as \( P_t^{a} \): 
\vspace{-2mm}
\begin{equation}
P_t^{a} = \bigcup_{i=1}^{N_t} \left(\mathbf{p}_i^{a}(t)\right).  
\vspace{-2mm}
\end{equation}

\textbf{Continuous 4D Point Cloud Fine Alignment.} Given that the 3D point clouds are estimated via single-step depth estimation and lack precise real-world values, aligning them based solely on parameters does not guarantee full alignment accuracy. Therefore, we introduce a fine alignment method, which refines the alignment over several iterations.

At each iteration \( k \), the rotation \( R_k \) and translation \( T_k \) are updated based on the point cloud alignment error \( \mathcal{E}_k \), which measures the difference between the transformed points and the reference alignment \( P_t^{r} \). The transformation parameters are updated by minimizing the alignment error:  
\vspace{-3mm}
\begin{equation}
\mathcal{E}_k = \sum_{i=1}^{N_t} \left\| \mathbf{p}_i^{a}(t) - \hat{\mathbf{p}}_i^{r}(t) \right\|_2^2 ,
\vspace{-3mm}
\end{equation} where \( \hat{\mathbf{p}}_i^{r}(t) \) represents the transformed 3D point \( \mathbf{p}_i \) after applying the current transformation parameters.

By applying this process iteratively to each frame, we generate a series of 3D point clouds \( P_{t_1}, P_{t_2}, \dots, P_{t_n} \). Finally, we obtain \( P_t^{\text{a}} \), the aligned point cloud, which supports subsequent spatial-temporal scene decoupling.

\begin{table*}[t!]
\centering
\caption{\textbf{Quantitative comparison of our model with the 3DGS method on both reconstruction and novel view synthesis (NVS).} Performance is evaluated using the Waymo-NOTR dataset \cite{yang2023emernerf}, with 'PSNR*' and 'SSIM*' referring to metrics for dynamic objects, where ENRF refers to EmerNeRF and $S^3$G refers to $S^3$Gaussian. The \textbf{best} results are highlighted in pink, and the \underline{second best} in blue.}
\vspace{-3mm}
\begin{tabular}{cc|ccccc|ccccccc}
\hline
\multirow{2}{*}{Data} & \multirow{2}{*}{Metrics} & \multicolumn{5}{c|}{Scene Reconstruction} & \multicolumn{5}{c}{Novel View Synthesis} \\
 & & 3DGS & MARS & ENRF & $S^3$G & Ours & 3DGS & MARS & ENRF & $S^3$G & Ours \\
\hline
\multirow{3}{*}{D32} & PSNR$\uparrow$ & 28.47 & 28.24 & 28.16 & \cellcolor{lightblue}31.35 & \cellcolor{lightpink}32.91 & 25.14 & 26.61 & 25.14 & \cellcolor{lightblue}27.44 & \cellcolor{lightpink}28.08 \\
 & SSIM$\uparrow$ & 0.876 & 0.866 & 0.806 & \cellcolor{lightblue}0.911 & \cellcolor{lightpink}0.934 & 0.813 & 0.796 & 0.747 & \cellcolor{lightblue}0.857 & \cellcolor{lightpink}0.864 \\
 & LPIPS$\downarrow$ & 0.136 & 0.252 & 0.228 & \cellcolor{lightblue}0.106 & \cellcolor{lightpink}0.104 & 0.165 & 0.305 & 0.313 & \cellcolor{lightblue}0.137 & \cellcolor{lightpink}0.129 \\
\hline
\multirow{2}{*}{D32} & PSNR*$\uparrow$ & 23.26 & 23.37 & 24.32 & \cellcolor{lightblue}26.02 & \cellcolor{lightpink}27.71 & 20.48 & 22.21 & \cellcolor{lightblue} 23.49 & 22.92 & \cellcolor{lightpink}24.98 \\
 & SSIM*$\uparrow$ & 0.716 & 0.701 & 0.682 & \cellcolor{lightblue}0.783 & \cellcolor{lightpink}0.796 & \cellcolor{lightpink}0.753 & 0.697 & 0.660 & \cellcolor{lightblue}0.680 & 0.644 \\
\hline
\multirow{3}{*}{S32} & PSNR$\uparrow$ & 29.42 & 28.31 & 30.00 & \cellcolor{lightblue}30.73 & \cellcolor{lightpink}31.03 & 26.82 & 27.63 & \cellcolor{lightpink}28.89 & 27.05 & \cellcolor{lightblue}28.56 \\
 & SSIM$\uparrow$ & 0.891 & 0.879 & 0.834 & \cellcolor{lightblue}0.883 & \cellcolor{lightpink}0.897 & 0.836 & 0.848 &  0.814 & \cellcolor{lightblue}0.825 & \cellcolor{lightpink}0.877 \\
 & LPIPS$\downarrow$ & 0.118 & 0.196 & 0.201 & \cellcolor{lightblue}0.116 & \cellcolor{lightpink}0.112 & 0.134 & 0.193 & 0.212 & \cellcolor{lightblue}0.142 & \cellcolor{lightpink} 0.132 \\
\hline
\label{exp1}
\end{tabular}
\vspace{-8mm}
\end{table*}

\subsection{Point-Conditioned Video Generation}
\label{sec:4D simulation}
Achieving Spatial-temporal decoupling is a critical aspect of autonomous driving simulation. However, existing models face challenges in separately capturing spatial and temporal variations within a scene due to limitations in their structure, making it difficult to decouple space and time within the same environment. To address this, Stag-1 processes a sequence of continuous 4D sparse point clouds \( P_t^{a} \) to generate 2D sparse keyframe videos \( V_{\text{key}}(t, \theta) \), under the dual control of \( C_p \) and \( T_t \). 
\vspace{-2mm}
\begin{equation}
 V_{key}(C_{p}, T_t) = \{f(P_t^{a} \mid C_p, T_t, t_k) \}, 
 \vspace{-2mm}
\end{equation} where, $t_k$ represents the discrete keyframe moments in the time sequence.

\textbf{Temporal Decoupling Keyframe.} Under a fixed camera pose, we proposed method for efficient spatial-temporal decoupling keyframe modeling by extracting the 3D point cloud for each keyframe and projecting it into a 2D image. Specifically, for each timestamp \( t_k \), we select the 3D point cloud corresponding to the current frame \( P_i(t_k) = \{(x, y, z) | (x, y, z) \in \mathbb{R}^3, t = t_k \} \), where \((x(t_k), y(t_k), z(t_k))\) represents the vehicle’s position at time \( t_k \), and \((r(t_k), i(t_k), j(t_k), k(t_k))\) represents the vehicle’s rotational quaternion at time \( t_k \), describing its orientation.
Subsequently, we project each keyframe's 3D point cloud \( P_i(t_k) \) using the camera’s matrix \( K \), along with the rotation matrix \( R(t_k) \) and translation vector \( t_k \) at timestamp \( t_k \), to obtain a sparse 2D point cloud in the image: 
\vspace{-1mm}
\begin{equation}
 V_{key}(t, \theta) =K \cdot \left[ R(t_k) | t_k \right] \cdot P_i(t_k) ,  P_i = C \end{equation}

Through this approach, dynamic point cloud data is accurately projected into a 2D image from a fixed viewpoint.

\textbf{Spatial Decoupling Keyframe.} In Spatial Decoupling Keyframe Modeling, we project the 3D point cloud of the current frame onto the 2D image plane using the aligned spatial information. Through perspective projection, we map the 3D point cloud to the 2D image plane: 
\vspace{-2mm}
\begin{equation}
  V_{key}(t, \theta) =K \cdot \left[ R(t_k) | t_k \right] \cdot P_i(t_k) ,  t_k = 0. 
  \vspace{-2mm}
\end{equation}

By using the aligned spatial information, we precisely convert the 3D point cloud of the current frame into its 2D projection. This method effectively leverages the spatial information, transforming it into a 2D point cloud representation while mitigating the impact of temporal variations on keyframe extraction.

\subsection{4D Spatial-Temporal Simulation }
\label{sec:a 4D Driving Simulation Model}

We present the overall training framework for our 4D generative simulation model in autonomous driving, as shown in Figure \ref{pic-main02-1}. First, we obtain \( P_t^{\text{a}} \) for all training scenes and align these within a 4D temporal context using \( C_p \) and \( T_t \) to iteratively refine the alignment. Our training follows a two-Stage approach: The time-focused Stage trains single-view scenes in a temporal context, while the spatial-focused Stage integrates surround-view information to capture spatial and temporal relationships.

\textbf{The Time-Focused Stage.} We use odd-frame sequential images \( I_{2n+1} \) (where \( n = 0, 1, 2, \dots \)) as ground truth and project even-frame 3D point clouds \( P_{2n+2} \) onto the image plane based on the poses \( T_{2n+1} \) and camera intrinsics \( K_{2n+1} \) of the odd frame: 
\vspace{-2mm}
\begin{equation}
  P_{2n+2}^{\text{proj}} = K_{2n+1} \cdot T_{2n+1} \cdot P_{2n+2} ,
  \vspace{-2mm}
\end{equation} where, \( P_{2n+2}^{\text{proj}} \) denotes the projection of even-frame point clouds using the parameters of the odd frames.

We generate paired training data by creating sequences of projected 3D point clouds \( P_{2n+2}^{\text{proj}} \) and their corresponding ground-truth images \( I_{2n+1} \). For training efficiency, we encode \( I_{2n+1} \) and the conditional signals \( P_{2n+2}^{\text{proj}} \) into latent space, where optimization is performed. To ensure accurate alignment and effective model learning, we define a custom loss function that guides the optimization process. The loss function is defined as follows: 
\vspace{-2mm}
\begin{equation}
 \min_{\phi} \mathbb{E}_{\tau \sim U(0,1), \eta \sim \mathcal{N}(0,I)} \left[ \| \eta_{\phi}(w_\tau, \tau, \hat{w}, I_{\text{ref}}) - \eta \|^2_2 \right],  
 \label{eq16} 
 \vspace{-1mm}
 \end{equation} where, \( \tau \) denotes the time step, sampled from a uniform distribution \( U(0,1) \);
\( \eta \) represents noise sampled from a standard normal distribution \( \mathcal{N}(0,I) \);
\( \phi \) represents the model parameters;
\( w_\tau \) represents the latent variable at time step \( \tau \);
\( \hat{w} \) is the conditional latent variable;
\( I_{\text{ref}} \) represents the reference image.

\textbf{The Spatial-Focused Stage.} We use the same input approach as in the time-focused Stage, with \( P_{2n+2}^{\text{proj}} \) as the sequence of 3D point cloud projections and \( I_{2n+1} \) as the reference images. To leverage overlapping information and interactions between surround-view images in autonomous driving, we introduce an attention mechanism for cross-image information exchange: 
\vspace{-2mm}
\begin{equation}
\mathrm{Att}\left( \cdot \right) =
\begin{cases}
\mathrm{S}\-\mathrm{A}( Q_{\mathrm{s}}, K_{\mathrm{s}}, V_{\mathrm{s}} ), &  Q_{\mathrm{s}}, K_{\mathrm{s}}, V_{\mathrm{s}} \in \mathbb{R}^{(B \times T) \times H \times W \times 6}, \\
\mathrm{T}\-\mathrm{A}( Q_{\mathrm{t}}, K_{\mathrm{t}}, V_{\mathrm{t}} ), &  Q_{\mathrm{t}}, K_{\mathrm{t}}, V_{\mathrm{t}} \in \mathbb{R}^{(B \times 6) \times H \times W \times T}.
\end{cases}
\vspace{-1mm}
\end{equation}
The proposed Stag-1 captures the spatial relationships within each frame across different perspectives and also considers temporal correlations between consecutive time steps. 
The training objective for this stage is designed similarly to \ref{eq16}.

\begin{figure}[t]
  \centering
  \includegraphics[width=1\linewidth]{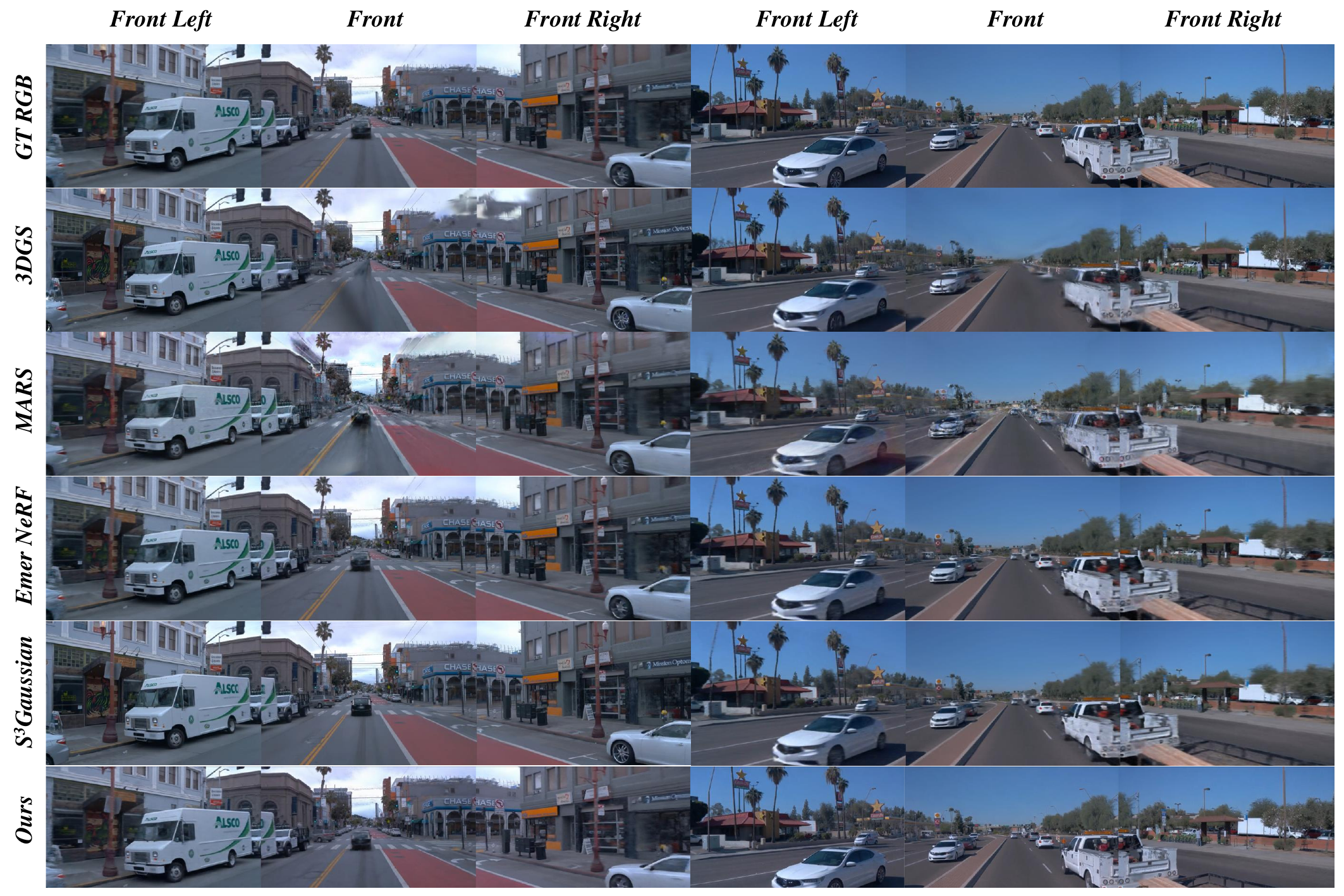}
    \vspace{-6mm}
  \caption{\textbf{Qualitative comparison on the Waymo-NOTR Datasets \cite{yang2023emernerf}.} Left shows novel view synthesis results, right shows dynamic scene reconstruction.}
  \label{test-pic02}
\vspace{-5mm}
\end{figure}

\begin{figure*}[t!]
  \centering
  \vspace{-8mm}
  \includegraphics[width=1\linewidth]{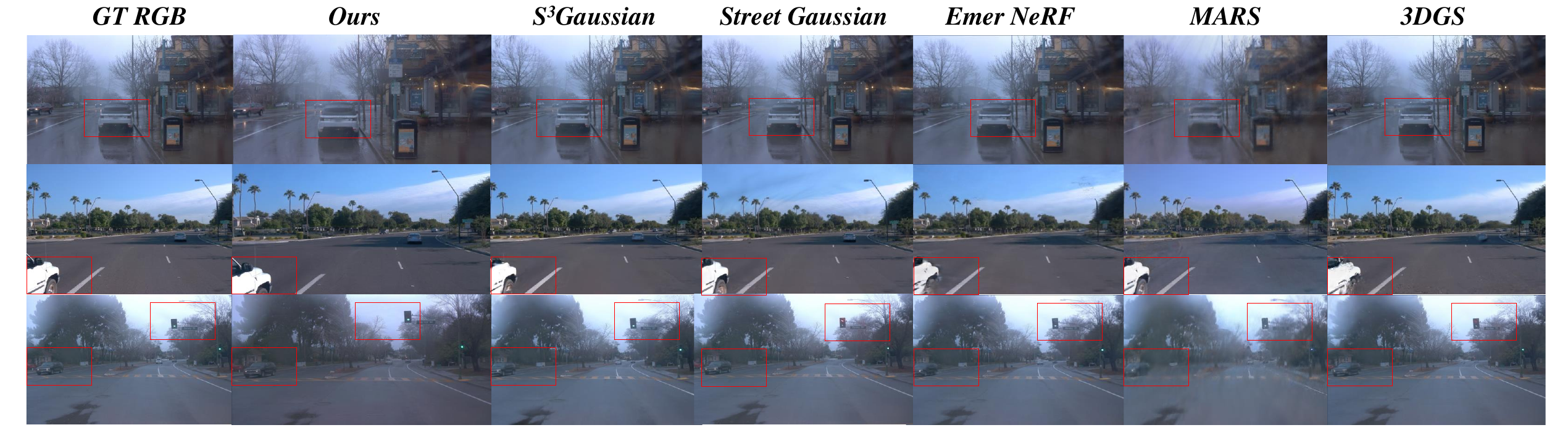}
    \vspace{-8mm}
  \caption{\textbf{Qualitative comparison on the Waymo-Street Datasets \cite{yan2024street}.} The results show that our method outperforms existing approaches in scene reconstruction. }
  \label{test-pic01}
\end{figure*}

\begin{table*}[t!]
\centering

\caption{\textbf{Quantitative comparison of our model with the 3DGS method on reconstruction in StreetGaussian datasets. \cite{yan2024street}} The results on the StreetGaussian dataset \cite{yan2024street} indicate that our proposed method outperforms related models across multiple evaluation metrics.}
\vspace{-3mm}
\begin{tabular}{c|cccccc|c}
\hline
Metrics & 3D GS & NSG & MARS & EmerNeRF & StreetGaussian & $S^3$Gaussian & Ours \\
\hline
PSNR$\uparrow$ & 29.64 & 28.31 & 31.37 & 32.34 & 34.96 & \cellcolor{lightblue}34.61 & \cellcolor{lightpink}35.23 \\
SSIM$\uparrow$ & 0.918 & 0.862 & 0.904 & 0.886 & 0.945 & \cellcolor{lightblue}0.950 & \cellcolor{lightpink}0.954 \\
LPIPS$\downarrow$ & 0.117 & 0.346 & 0.246 & 0.142 & 0.068 & \cellcolor{lightblue}0.05 & \cellcolor{lightpink}0.043 \\
PSNR*$\uparrow$ & 16.48 & 19.55 & 23.07 & 25.71 & 25.46 & \cellcolor{lightblue}25.78 & \cellcolor{lightpink}26.42 \\
\hline
\end{tabular}
\vspace{-3mm}
\label{exp3}
\end{table*}

\begin{table*}[t!]
\centering
\caption{\textbf{Quantitative comparison of our model with the 3DGS method on reconstruction in Waymo datasets. \cite{sun2020scalability}.} Overall performance of our methods compared to existing approaches on the Waymo Open Datasets (WOD) \cite{sun2020scalability}.}
   \vspace{-3mm}
\begin{tabular}{cc|ccccc|ccccc}
\hline

\multirow{3}{*}{Data} & \multirow{3}{*}{Metrics} & \multicolumn{5}{c|}{Front View} & \multicolumn{5}{c}{Multi-view} \\

 & & \multirow{2}{*}{3DGS} & Street& Emer & \multirow{2}{*}{FreeVS} & \multirow{2}{*}{Ours}  & \multirow{2}{*}{3DGS} & Street& Emer & \multirow{2}{*}{FreeVS} & \multirow{2}{*}{Ours}  \\
 & & &  Gaussian & NeRF & & & &  Gaussian & NeRF & & \\
\hline

\multirow{3}{*}{WOD} & PSNR$\uparrow$ & 26.31 & \cellcolor{lightblue}30.80 & 30.28 & 25.30& \cellcolor{lightpink}31.23 & 19.21 &  22.47 & 24.68 & \cellcolor{lightblue}24.96 & \cellcolor{lightpink}26.03 \\
 & SSIM$\uparrow$ & 0.799 & \cellcolor{lightblue}0.903 & 0.869 & 0.787 & \cellcolor{lightpink}0.912 & 0.586 & 0.702 &  0.689 & \cellcolor{lightblue}0.730 & \cellcolor{lightpink}0.751 \\
 & LPIPS$\downarrow$ & 0.143 & \cellcolor{lightpink}0.096 & 0.155 & 0.139 & \cellcolor{lightblue}0.127 & 0.366 & 0.314 &  0.347 & \cellcolor{lightblue}0.179 & \cellcolor{lightpink}0.141 \\
\hline
\label{exp2}
\end{tabular}
\vspace{-10mm}
\end{table*}

\begin{figure*}[t!]
     \vspace{-4mm}
  \centering

  \includegraphics[width=1\linewidth]{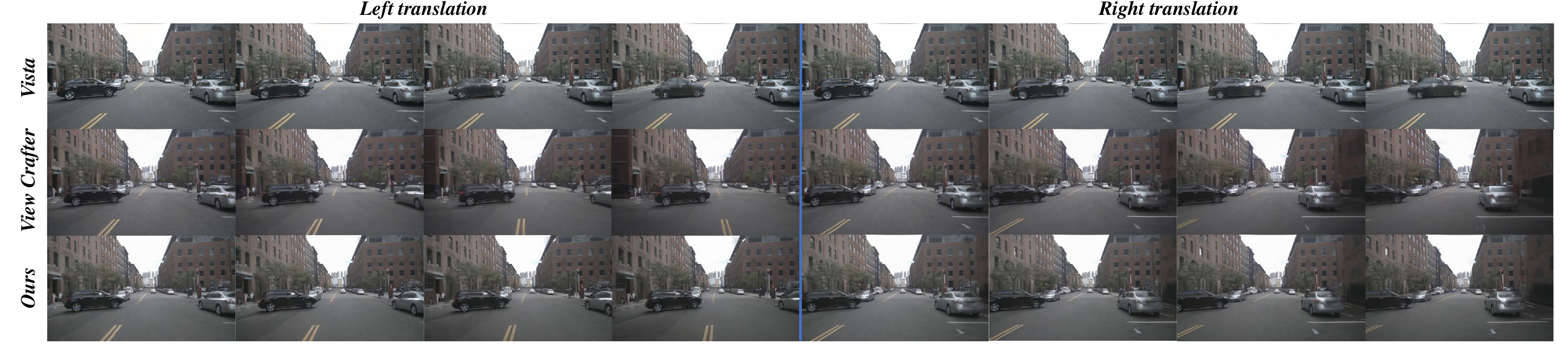}
      \vspace{-8mm}
  \caption{\textbf{We compared viewpoint translation under frozen time conditions on the NuScenes dataset \cite{caesar2020nuscenes} .} Our method successfully applied left and right viewpoint translations to the camera pose.}
     \vspace{-1mm}
  \label{test-pic03}
    \vspace{-3mm}
\end{figure*}

\begin{figure*}[t!]
  \centering
  \includegraphics[width=1\linewidth]{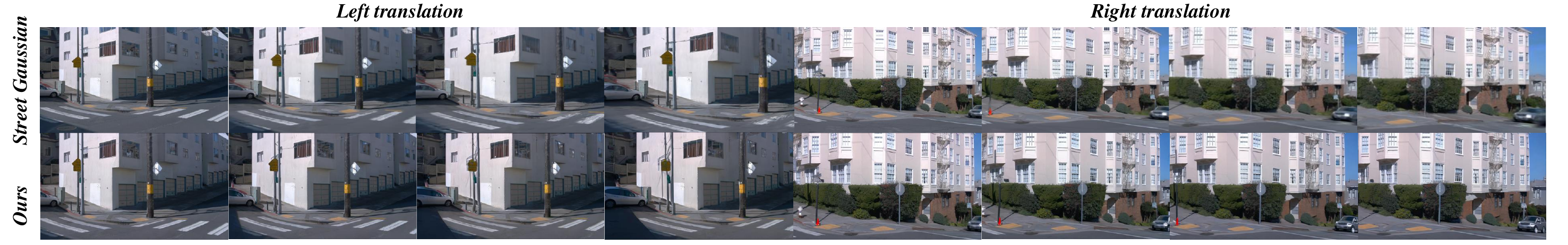}
    \vspace{-8mm}
  \caption{\textbf{We compared viewpoint translation under frozen time conditions on the Waymo dataset \cite{sun2020scalability}. } Our method effectively performed left and right viewpoint translations on the camera pose.}
    \vspace{-4mm}
  \label{test-pic03.5}
  \vspace{-0.5mm}
\end{figure*}

\vspace{3mm}
\section{Experiments}
In this section, we describe our training methodology and evaluate the effectiveness of Stag-1 for autonomous driving tasks, specifically focusing on 3D reconstruction and 4D simulation. Quantitative results from comparison and ablation studies assess Stag-1 performance, with visualizations of scene reconstruction, viewpoint translation, and static temporal conditions. The structure of our experimental results is as follows:

\subsection{Implementation Details}

Our training process consists of two Stages. In the first Stage, we pre-train the model following \cite{yu2024viewcrafter} for 5,000 iterations. Using sweep records from the NuScenes dataset \cite{caesar2020nuscenes} , we randomly select starting points and sample eight consecutive frames, where four frames are used as conditions and four as ground truth. During this Stage, we freeze the Encoder module and train only the Decoder module using the AdamW optimizer \cite{loshchilov2017decoupled} with a learning rate of \(1 \times 10^{-5}\). In the second Stage, to further learn spatial relationships across panoramic views, we freeze all remaining components and train only the spatial attention module for an additional 3,000 iterations, again using the AdamW optimizer \cite{loshchilov2017decoupled} with a learning rate of \(1 \times 10^{-5}\).

\subsection{4D Reconstruction and Synthesis}

To assess our method's capabilities in 4D reconstruction, we conduct zero-shot evaluations on the Waymo-NOTR dataset \cite{yang2023emernerf}, as shown in Table \ref{exp1}. Our approach demonstrates superior performance in scene reconstruction and novel view synthesis compared to existing methods \cite{kerbl20233d, wu2023mars, yang2023emernerf, yan2024street, huang2024textit}. For the static-32 dataset \cite{yan2024street}, we follow conventional metrics \cite{yang2023emernerf} using PSNR, SSIM, and LPIPS \cite{zhang2018unreasonable} to evaluate rendering quality, and for dynamic data, we use PSNR* and SSIM* \cite{huang2024textit} to focus on dynamic objects. Our results outperform other methods, showcasing the model's generalization under zero-shot conditions and its ability to model both static scenes and dynamic objects. Qualitatively, as shown in Figure \ref{test-pic02}, our method excels in monocular scene reconstruction and multi-view synthesis. Additionally, we conducted both quantitative and qualitative evaluations of scene reconstruction on the Street Gaussian dataset \cite{yan2024street}, as presented in Table \ref{exp3}, with visualizations shown in Figure \ref{test-pic01}.

Furthermore, recent studies \cite{wang2024freevs, yang2023emernerf, yan2024street} have used the Waymo Open Dataset (WOD) \cite{sun2020scalability} for evaluations. To accurately compare our method with the latest approaches, we conducted quantitative analyses under similar experimental conditions. As shown in Table \ref{exp2}, our method outperforms other approaches in reconstruction. Thus, the quantitative comparisons and visualization results across the three different experimental conditions demonstrate that our proposed reconstruction and novel view synthesis methods outperform other related approaches.

\subsection{4D Driving Simulation }

Autonomous driving generative 4D simulation based on real-world scenes requires the ability to decouple spatial-temporal relationships. This involves observing the scene from different camera viewpoints based on the current time state or decomposing temporal motion based on a fixed spatial state. We conducted both quantitative and qualitative comparison experiments on the NuScenes \cite{caesar2020nuscenes} and Waymo \cite{sun2020scalability} datasets to demonstrate the capability and effectiveness of the proposed method.

\begin{table}[t!]
\centering
\setlength{\tabcolsep}{4.5pt}
\begin{tabular}{lcccccccccc}
\hline
\textbf{Methods} & \multicolumn{2}{c}{\textbf{Frozen time}} & \multicolumn{2}{c}{\textbf{Frozen space}} & \multicolumn{2}{c}{\textbf{Remove cars}} \\
 & FID$\downarrow$ & FVD$\downarrow$  & FID$\downarrow$ & FVD$\downarrow$  & FID$\downarrow$ & FVD$\downarrow$  \\
\hline
Vista & 89.7 & 356.8  & 94.6 & 473.2  & 92.5 & 492.7 \\
VC & 71.8 & 218.6 & 89.2 & 314.4  & 84.7 & 223.8 \\
Ours & 34.9 & 91.5  & 28.3 & 84.4  & 47.1 & 127.6  \\
\hline

\end{tabular}
  \vspace{-3mm}
\caption{\textbf{Performance comparison under different conditions.} We evaluated the quality of image and video generation under various controls, and our model outperforms the others. Here, VC stands for ViewCrafter.}
\label{exp5}
  \vspace{-5mm}
\end{table}

\begin{figure*}[t!]
  \centering
    \vspace{-3mm}
  \includegraphics[width=1\linewidth]{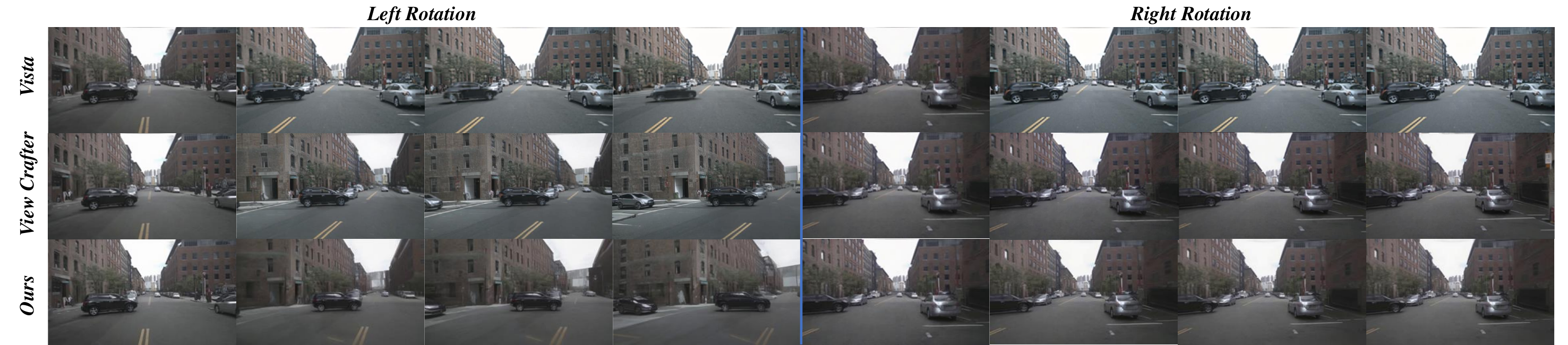}
      \vspace{-7mm}
  \caption{\textbf{We compared viewpoint rotation under frozen time conditions on the NuScenes dataset.\cite{caesar2020nuscenes} } Our method applied left and right viewpoint rotations to the camera pose and successfully achieved the desired transformations. }
  \label{test-pic04}
\vspace{-5mm}
\end{figure*}

\vspace{0.48mm}
\textbf{Frozen Time.} A key aspect of autonomous driving 4D simulation is the ability to achieve dynamic viewpoint changes under frozen temporal conditions. We compared our proposed method with existing approaches, and the visualization results show that our method successfully achieves the desired tasks. As shown in Figure \ref{test-pic03}, we translated the camera pose in the NuScenes dataset \cite{caesar2020nuscenes}. The results indicate that our method can achieve accurate translations. To provide a fair comparison with 3DGS-based methods, we conducted similar tests on the Waymo dataset \cite{sun2020scalability}. As shown in Figure \ref{test-pic03.5}, our method outperforms others in terms of image accuracy. Additionally, we performed viewpoint rotation to test the model's ability to handle diverse camera transformations. As shown in Figure \ref{test-pic04}, our model successfully accomplished this task.

\vspace{-3mm}
\begin{figure*}[t!]
  \centering
  \includegraphics[width=1\linewidth]{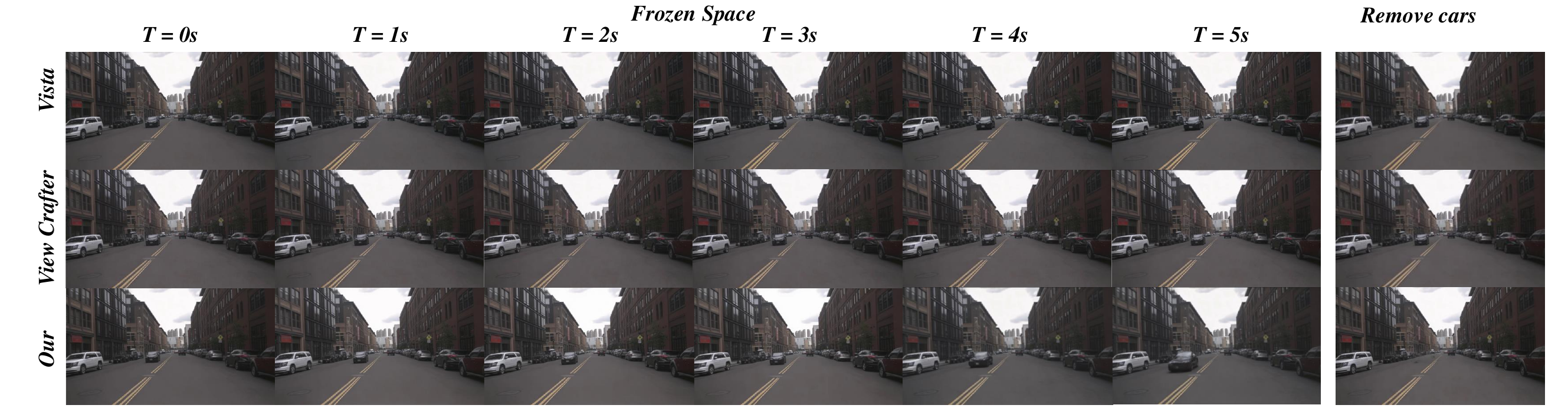}
      \vspace{-8mm}
  \caption{\textbf{In the frozen viewpoint scenario, we fixed the camera position to apply temporal transformations.} As shown on the left, our model accurately represents realistic vehicle motion under spatially stationary conditions.}
  \label{test-pic05}
\vspace{-5mm}
\end{figure*}

\vspace{1.5mm}
\textbf{Frozen Space.} Another key aspect of 4D simulation is its ability to vary temporal ranges while keeping the camera position fixed, enabling diverse functionalities. We demonstrate this capability by simulating temporal movement under frozen spatial conditions, as shown in the left of Figure \ref{test-pic05}. The figure illustrates moving vehicles relative to the ground truth, while the background remains stable, validating the method’s ability to simulate time variations in a fixed spatial context. To quantitatively assess image and video quality under viewpoint translation and transformation, we use the Fréchet Inception Distance (FID) \cite{heusel2017gans} for images and Fréchet Video Distance (FVD) for videos. As shown in Table \ref{exp5}, the quantitative results demonstrate that our method outperforms previous approaches and effectively decouples spatial and temporal relationships.

\vspace{3mm}
\textbf{Multi View Simulation.} We propose a method for 4D surround-view simulation in autonomous driving, capable of generating dynamic images with consistent transformations across different settings, as demonstrated in the lower part of Figure \ref{pic-main00}. This method significantly improves interactive simulations that rely on surround-view information.

\textbf{Remove Cars.} Furthermore, we showcase the enhanced capabilities of our proposed model. After aligning the 4D point cloud scene, we can selectively remove specific point clouds to eliminate individual vehicles, as illustrated on the right side of Figure \ref{test-pic05}.

\subsection{Ablation Studies}

\begin{figure}[t]
  \centering
  \vspace{-2mm}
  \includegraphics[width=1\linewidth]{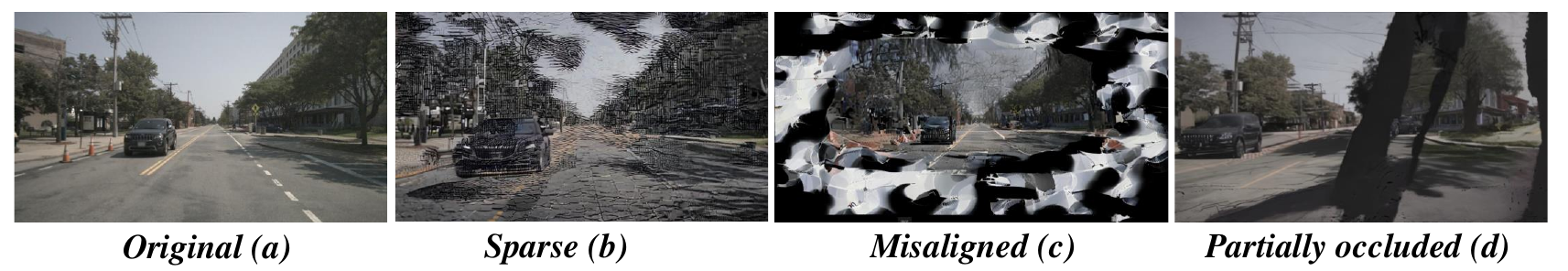}
  \vspace{-5mm}
  \caption{\textbf{Ablation study under varying conditions.} We evaluate the performance under different conditions: (a) the original scene image; (b) simulation results with sparse point clouds; (c) simulation results with precisely aligned point clouds; and (d) performance with partial point cloud data missing.}
  \label{test-pic07}
  \vspace{-8mm}
\end{figure}

To assess the impact of different point cloud conditions on simulation quality, we conducted ablation studies, as shown in Figure \ref{test-pic07}.

\textbf{Sparse Point Cloud Information.} Point cloud density directly affects reconstruction quality. Sparse point clouds result in incomplete scenes, leading to noise and incorrect information in the output, as seen in Figure \ref{test-pic07} (b). \textbf{Misaligned 4D Point Clouds.} Misalignment of point clouds with coordinate parameters causes the loss of keyframe information. As shown in Figure \ref{test-pic07} (c), misaligned point clouds introduce significant empty pixels, which causes confusion during model completion and results in blurry or disordered scenes. \textbf{Partial Occlusion in Panoramic Point Clouds.} In cases of occlusion, the model attempts to fill in missing data, as shown in Figure \ref{test-pic07} (d). However, accurately reconstructing occluded regions remains challenging, hindering precise simulation in these areas. These ablation studies highlight the importance of each component in improving model performance.

  \vspace{-8mm}

  \vspace{3mm}
\section{Conclusion}

In this paper, we propose a generative 4D simulation model for autonomous driving, designed to edit real-world scenes for controllable autonomous driving simulation. We reconstruct coherent and aligned real-world scenes in 4D point clouds and design keyframe projections to decouple spatial and temporal relationships. Finally, we build a generative network for 4D simulation on sparse point clouds. Both visualization and quantitative results show that the proposed method can extract key elements of real scenes for controllable simulation, providing a feasible solution for autonomous driving testing and validation.

\textbf{Limitations.} Controlling vehicle or pedestrian motion via rigid body clustering of point clouds to enhance scene editing is crucial for simulation tasks and will be addressed in future research.


\begin{thebibliography}{58}
\providecommand{\natexlab}[1]{#1}
\providecommand{\url}[1]{\texttt{#1}}
\expandafter\ifx\csname urlstyle\endcsname\relax
  \providecommand{\doi}[1]{doi: #1}\else
  \providecommand{\doi}{doi: \begingroup \urlstyle{rm}\Url}\fi

\bibitem[Blattmann et~al.(2023)Blattmann, Dockhorn, Kulal, Mendelevitch, Kilian, Lorenz, Levi, English, Voleti, Letts, et~al.]{blattmann2023stable}
Andreas Blattmann, Tim Dockhorn, Sumith Kulal, Daniel Mendelevitch, Maciej Kilian, Dominik Lorenz, Yam Levi, Zion English, Vikram Voleti, Adam Letts, et~al.
\newblock Stable video diffusion: Scaling latent video diffusion models to large datasets.
\newblock \emph{arXiv preprint arXiv:2311.15127}, 2023.

\bibitem[Caesar et~al.(2020)Caesar, Bankiti, Lang, Vora, Liong, Xu, Krishnan, Pan, Baldan, and Beijbom]{caesar2020nuscenes}
Holger Caesar, Varun Bankiti, Alex~H Lang, Sourabh Vora, Venice~Erin Liong, Qiang Xu, Anush Krishnan, Yu Pan, Giancarlo Baldan, and Oscar Beijbom.
\newblock nuscenes: A multimodal dataset for autonomous driving.
\newblock In \emph{CVPR}, pages 11621--11631, 2020.

\bibitem[Caesar et~al.(2021)Caesar, Kabzan, Tan, Fong, Wolff, Lang, Fletcher, Beijbom, and Omari]{caesar2021nuplan}
Holger Caesar, Juraj Kabzan, Kok~Seang Tan, Whye~Kit Fong, Eric Wolff, Alex Lang, Luke Fletcher, Oscar Beijbom, and Sammy Omari.
\newblock nuplan: A closed-loop ml-based planning benchmark for autonomous vehicles.
\newblock \emph{arXiv preprint arXiv:2106.11810}, 2021.

\bibitem[Ceylan et~al.(2023)Ceylan, Huang, and Mitra]{ceylan2023pix2video}
Duygu Ceylan, Chun-Hao~P Huang, and Niloy~J Mitra.
\newblock Pix2video: Video editing using image diffusion.
\newblock In \emph{ICCV}, pages 23206--23217, 2023.

\bibitem[Chen et~al.(2024)Chen, Zhang, Xie, Li, Zhang, Lu, and Zhang]{chen2024s}
Yurui Chen, Junge Zhang, Ziyang Xie, Wenye Li, Feihu Zhang, Jiachen Lu, and Li Zhang.
\newblock S-nerf++: Autonomous driving simulation via neural reconstruction and generation.
\newblock \emph{arXiv preprint arXiv:2402.02112}, 2024.

\bibitem[Croitoru et~al.(2023)Croitoru, Hondru, Ionescu, and Shah]{croitoru2023diffusion}
Florinel-Alin Croitoru, Vlad Hondru, Radu~Tudor Ionescu, and Mubarak Shah.
\newblock Diffusion models in vision: A survey.
\newblock \emph{IEEE Transactions on Pattern Analysis and Machine Intelligence}, 45\penalty0 (9):\penalty0 10850--10869, 2023.

\bibitem[Dosovitskiy et~al.(2017)Dosovitskiy, Ros, Codevilla, Lopez, and Koltun]{dosovitskiy2017carla}
Alexey Dosovitskiy, German Ros, Felipe Codevilla, Antonio Lopez, and Vladlen Koltun.
\newblock Carla: An open urban driving simulator.
\newblock In \emph{Conference on robot learning}, pages 1--16. PMLR, 2017.

\bibitem[Feng et~al.(2023)Feng, Sun, Yan, Zhu, Zou, Shen, and Liu]{feng2023dense}
Shuo Feng, Haowei Sun, Xintao Yan, Haojie Zhu, Zhengxia Zou, Shengyin Shen, and Henry~X Liu.
\newblock Dense reinforcement learning for safety validation of autonomous vehicles.
\newblock \emph{Nature}, 615\penalty0 (7953):\penalty0 620--627, 2023.

\bibitem[Gao et~al.(2020)Gao, Sun, Zhao, Shen, Anguelov, Li, and Schmid]{gao2020vectornet}
Jiyang Gao, Chen Sun, Hang Zhao, Yi Shen, Dragomir Anguelov, Congcong Li, and Cordelia Schmid.
\newblock Vectornet: Encoding hd maps and agent dynamics from vectorized representation.
\newblock In \emph{CVPR}, pages 11525--11533, 2020.

\bibitem[Gao et~al.(2023)Gao, Chen, Xie, Hong, Li, Yeung, and Xu]{gao2023magicdrive}
Ruiyuan Gao, Kai Chen, Enze Xie, Lanqing Hong, Zhenguo Li, Dit-Yan Yeung, and Qiang Xu.
\newblock Magicdrive: Street view generation with diverse 3d geometry control.
\newblock \emph{arXiv preprint arXiv:2310.02601}, 2023.

\bibitem[Gao et~al.(2024{\natexlab{a}})Gao, Chen, Li, Hong, Li, and Xu]{gao2024magicdrive3d}
Ruiyuan Gao, Kai Chen, Zhihao Li, Lanqing Hong, Zhenguo Li, and Qiang Xu.
\newblock Magicdrive3d: Controllable 3d generation for any-view rendering in street scenes.
\newblock \emph{arXiv preprint arXiv:2405.14475}, 2024{\natexlab{a}}.

\bibitem[Gao et~al.(2024{\natexlab{b}})Gao, Yang, Chen, Chitta, Qiu, Geiger, Zhang, and Li]{gao2024vista}
Shenyuan Gao, Jiazhi Yang, Li Chen, Kashyap Chitta, Yihang Qiu, Andreas Geiger, Jun Zhang, and Hongyang Li.
\newblock Vista: A generalizable driving world model with high fidelity and versatile controllability.
\newblock \emph{arXiv preprint arXiv:2405.17398}, 2024{\natexlab{b}}.

\bibitem[Giannone et~al.(2023)Giannone, Srivastava, Winther, and Ahmed]{giannone2023aligning}
Giorgio Giannone, Akash Srivastava, Ole Winther, and Faez Ahmed.
\newblock Aligning optimization trajectories with diffusion models for constrained design generation.
\newblock \emph{NeurIPS}, 36:\penalty0 51830--51861, 2023.

\bibitem[Goodfellow et~al.(2014)Goodfellow, Pouget-Abadie, Mirza, Xu, Warde-Farley, Ozair, Courville, and Bengio]{goodfellow2014generative}
Ian Goodfellow, Jean Pouget-Abadie, Mehdi Mirza, Bing Xu, David Warde-Farley, Sherjil Ozair, Aaron Courville, and Yoshua Bengio.
\newblock Generative adversarial nets.
\newblock \emph{NeurIPS}, 27, 2014.

\bibitem[Heusel et~al.(2017)Heusel, Ramsauer, Unterthiner, Nessler, and Hochreiter]{heusel2017gans}
Martin Heusel, Hubert Ramsauer, Thomas Unterthiner, Bernhard Nessler, and Sepp Hochreiter.
\newblock Gans trained by a two time-scale update rule converge to a local nash equilibrium.
\newblock \emph{NeurIPS}, 30, 2017.

\bibitem[Hu et~al.(2023{\natexlab{a}})Hu, Russell, Yeo, Murez, Fedoseev, Kendall, Shotton, and Corrado]{hu2023gaia}
Anthony Hu, Lloyd Russell, Hudson Yeo, Zak Murez, George Fedoseev, Alex Kendall, Jamie Shotton, and Gianluca Corrado.
\newblock Gaia-1: A generative world model for autonomous driving.
\newblock \emph{arXiv preprint arXiv:2309.17080}, 2023{\natexlab{a}}.

\bibitem[Hu et~al.(2022)Hu, Chen, Wu, Li, Yan, and Tao]{hu2022stp3}
Shengchao Hu, Li Chen, Penghao Wu, Hongyang Li, Junchi Yan, and Dacheng Tao.
\newblock St-p3: End-to-end vision-based autonomous driving via spatial-temporal feature learning.
\newblock In \emph{ECCV}, 2022.

\bibitem[Hu et~al.(2023{\natexlab{b}})Hu, Yang, Chen, Li, Sima, Zhu, Chai, Du, Lin, Wang, et~al.]{hu2023planning}
Yihan Hu, Jiazhi Yang, Li Chen, Keyu Li, Chonghao Sima, Xizhou Zhu, Siqi Chai, Senyao Du, Tianwei Lin, Wenhai Wang, et~al.
\newblock Planning-oriented autonomous driving.
\newblock In \emph{CVPR}, pages 17853--17862, 2023{\natexlab{b}}.

\bibitem[Huang et~al.(2024)Huang, Wei, Zheng, An, Lu, Zhan, Tomizuka, Keutzer, and Zhang]{huang2024textit}
Nan Huang, Xiaobao Wei, Wenzhao Zheng, Pengju An, Ming Lu, Wei Zhan, Masayoshi Tomizuka, Kurt Keutzer, and Shanghang Zhang.
\newblock $s^3$gaussian: Self-supervised street gaussians for autonomous driving.
\newblock \emph{arXiv preprint arXiv:2405.20323}, 2024.

\bibitem[Jiang et~al.(2023)Jiang, Chen, Xu, Liao, Chen, Zhou, Zhang, Liu, Huang, and Wang]{jiang2023vad}
Bo Jiang, Shaoyu Chen, Qing Xu, Bencheng Liao, Jiajie Chen, Helong Zhou, Qian Zhang, Wenyu Liu, Chang Huang, and Xinggang Wang.
\newblock Vad: Vectorized scene representation for efficient autonomous driving.
\newblock \emph{ICCV}, 2023.

\bibitem[Karras et~al.(2023)Karras, Holynski, Wang, and Kemelmacher-Shlizerman]{karras2023dreampose}
Johanna Karras, Aleksander Holynski, Ting-Chun Wang, and Ira Kemelmacher-Shlizerman.
\newblock Dreampose: Fashion image-to-video synthesis via stable diffusion.
\newblock In \emph{ICCV}, pages 22623--22633. IEEE, 2023.

\bibitem[Kerbl et~al.(2023)Kerbl, Kopanas, Leimk{\"u}hler, and Drettakis]{kerbl20233d}
Bernhard Kerbl, Georgios Kopanas, Thomas Leimk{\"u}hler, and George Drettakis.
\newblock 3d gaussian splatting for real-time radiance field rendering.
\newblock \emph{ACM Trans. Graph.}, 42\penalty0 (4):\penalty0 139--1, 2023.

\bibitem[Kim et~al.(2021)Kim, Philion, Torralba, and Fidler]{kim2021drivegan}
Seung~Wook Kim, Jonah Philion, Antonio Torralba, and Sanja Fidler.
\newblock Drivegan: Towards a controllable high-quality neural simulation.
\newblock In \emph{CVPR}, pages 5820--5829, 2021.

\bibitem[Li et~al.(2019)Li, Qi, Lukasiewicz, and Torr]{li2019controllable}
Bowen Li, Xiaojuan Qi, Thomas Lukasiewicz, and Philip Torr.
\newblock Controllable text-to-image generation.
\newblock \emph{NeurIPS}, 32, 2019.

\bibitem[Li et~al.(2025)Li, Yang, Kuang, Wu, Wang, Xiao, and Chen]{li2025controlnet}
Ming Li, Taojiannan Yang, Huafeng Kuang, Jie Wu, Zhaoning Wang, Xuefeng Xiao, and Chen Chen.
\newblock Controlnet $++$: Improving conditional controls with efficient consistency feedback.
\newblock In \emph{ECCV}, pages 129--147. Springer, 2025.

\bibitem[Li et~al.(2022)Li, Wang, Li, Xie, Sima, Lu, Qiao, and Dai]{li2022bevformer}
Zhiqi Li, Wenhai Wang, Hongyang Li, Enze Xie, Chonghao Sima, Tong Lu, Yu Qiao, and Jifeng Dai.
\newblock Bevformer: Learning bird’s-eye-view representation from multi-camera images via spatiotemporal transformers.
\newblock In \emph{ECCV}, pages 1--18. Springer, 2022.

\bibitem[Lindstr{\"o}m et~al.(2024)Lindstr{\"o}m, Hess, Lilja, Fatemi, Hammarstrand, Petersson, and Svensson]{lindstrom2024nerfs}
Carl Lindstr{\"o}m, Georg Hess, Adam Lilja, Maryam Fatemi, Lars Hammarstrand, Christoffer Petersson, and Lennart Svensson.
\newblock Are nerfs ready for autonomous driving? towards closing the real-to-simulation gap.
\newblock In \emph{CVPR}, pages 4461--4471, 2024.

\bibitem[Liu et~al.(2024{\natexlab{a}})Liu, Gong, Zhang, Lu, Zhou, and Liao]{liu2024glmdrivenet}
Wenzhuo Liu, Yan Gong, Guoying Zhang, Jianli Lu, Yunlai Zhou, and Junbin Liao.
\newblock Glmdrivenet: Global--local multimodal fusion driving behavior classification network.
\newblock \emph{Engineering Applications of Artificial Intelligence}, 129:\penalty0 107575, 2024{\natexlab{a}}.

\bibitem[Liu et~al.(2024{\natexlab{b}})Liu, Lu, Liao, Qiao, Zhang, Zhu, Xu, and Li]{liu2024fmdnet}
Wenzhuo Liu, Jianli Lu, Junbin Liao, Yicheng Qiao, Guoying Zhang, Jiayin Zhu, Bozhang Xu, and Zhiwei Li.
\newblock Fmdnet: Feature-attention-embedding-based multimodal-fusion driving-behavior-classification network.
\newblock \emph{IEEE Transactions on Computational Social Systems}, 2024{\natexlab{b}}.

\bibitem[Liu et~al.(2023)Liu, Yan, Jia, Li, Gao, Wang, and Zhang]{liu2023petrv2}
Yingfei Liu, Junjie Yan, Fan Jia, Shuailin Li, Aqi Gao, Tiancai Wang, and Xiangyu Zhang.
\newblock Petrv2: A unified framework for 3d perception from multi-camera images.
\newblock In \emph{ICCV}, pages 3262--3272, 2023.

\bibitem[Loshchilov(2017)]{loshchilov2017decoupled}
I Loshchilov.
\newblock Decoupled weight decay regularization.
\newblock \emph{arXiv preprint arXiv:1711.05101}, 2017.

\bibitem[Lu et~al.(2025)Lu, Huang, Yang, Zhang, and Zhang]{lu2025wovogen}
Jiachen Lu, Ze Huang, Zeyu Yang, Jiahui Zhang, and Li Zhang.
\newblock Wovogen: World volume-aware diffusion for controllable multi-camera driving scene generation.
\newblock In \emph{ECCV}, pages 329--345. Springer, 2025.

\bibitem[Mildenhall et~al.(2021)Mildenhall, Srinivasan, Tancik, Barron, Ramamoorthi, and Ng]{mildenhall2021nerf}
Ben Mildenhall, Pratul~P Srinivasan, Matthew Tancik, Jonathan~T Barron, Ravi Ramamoorthi, and Ren Ng.
\newblock Nerf: Representing scenes as neural radiance fields for view synthesis.
\newblock \emph{Communications of the ACM}, 65\penalty0 (1):\penalty0 99--106, 2021.

\bibitem[Peebles and Xie(2023)]{peebles2023scalable}
William Peebles and Saining Xie.
\newblock Scalable diffusion models with transformers.
\newblock In \emph{ICCV}, pages 4195--4205, 2023.

\bibitem[Ramesh et~al.(2021)Ramesh, Pavlov, Goh, Gray, Voss, Radford, Chen, and Sutskever]{ramesh2021zero}
Aditya Ramesh, Mikhail Pavlov, Gabriel Goh, Scott Gray, Chelsea Voss, Alec Radford, Mark Chen, and Ilya Sutskever.
\newblock Zero-shot text-to-image generation.
\newblock In \emph{ICML}, pages 8821--8831. Pmlr, 2021.

\bibitem[Rombach et~al.(2022)Rombach, Blattmann, Lorenz, Esser, and Ommer]{rombach2022high}
Robin Rombach, Andreas Blattmann, Dominik Lorenz, Patrick Esser, and Bj{\"o}rn Ommer.
\newblock High-resolution image synthesis with latent diffusion models.
\newblock In \emph{CVPR}, pages 10684--10695, 2022.

\bibitem[Schmied et~al.(2023)Schmied, Fischer, Danelljan, Pollefeys, and Yu]{schmied2023r3d3}
Aron Schmied, Tobias Fischer, Martin Danelljan, Marc Pollefeys, and Fisher Yu.
\newblock R3d3: Dense 3d reconstruction of dynamic scenes from multiple cameras.
\newblock In \emph{ICCV}, pages 3216--3226, 2023.

\bibitem[Shao et~al.(2024)Shao, Hu, Wang, Song, Waslander, Liu, and Li]{shao2024lmdrive}
Hao Shao, Yuxuan Hu, Letian Wang, Guanglu Song, Steven~L Waslander, Yu Liu, and Hongsheng Li.
\newblock Lmdrive: Closed-loop end-to-end driving with large language models.
\newblock In \emph{CVPR}, pages 15120--15130, 2024.

\bibitem[Sun et~al.(2020)Sun, Kretzschmar, Dotiwalla, Chouard, Patnaik, Tsui, Guo, Zhou, Chai, Caine, et~al.]{sun2020scalability}
Pei Sun, Henrik Kretzschmar, Xerxes Dotiwalla, Aurelien Chouard, Vijaysai Patnaik, Paul Tsui, James Guo, Yin Zhou, Yuning Chai, Benjamin Caine, et~al.
\newblock Scalability in perception for autonomous driving: Waymo open dataset.
\newblock In \emph{CVPR}, pages 2446--2454, 2020.

\bibitem[Tonderski et~al.(2024)Tonderski, Lindstr{\"o}m, Hess, Ljungbergh, Svensson, and Petersson]{tonderski2024neurad}
Adam Tonderski, Carl Lindstr{\"o}m, Georg Hess, William Ljungbergh, Lennart Svensson, and Christoffer Petersson.
\newblock Neurad: Neural rendering for autonomous driving.
\newblock In \emph{CVPR}, pages 14895--14904, 2024.

\bibitem[Wang et~al.(2021)Wang, Liu, and Hsu]{wang2021simulation}
Chia-Sui Wang, Ding-Yu Liu, and Kuei-Shu Hsu.
\newblock Simulation and application of cooperative driving sense systems using prescan software.
\newblock \emph{Microsystem Technologies}, 27\penalty0 (4):\penalty0 1201--1210, 2021.

\bibitem[Wang et~al.(2024)Wang, Fan, Wang, Chen, and Zhang]{wang2024freevs}
Qitai Wang, Lue Fan, Yuqi Wang, Yuntao Chen, and Zhaoxiang Zhang.
\newblock Freevs: Generative view synthesis on free driving trajectory.
\newblock \emph{arXiv preprint arXiv:2410.18079}, 2024.

\bibitem[Wang et~al.(2023)Wang, Zhu, Huang, Chen, Zhu, and Lu]{wang2023drivedreamer}
Xiaofeng Wang, Zheng Zhu, Guan Huang, Xinze Chen, Jiagang Zhu, and Jiwen Lu.
\newblock Drivedreamer: Towards real-world-driven world models for autonomous driving.
\newblock \emph{arXiv preprint arXiv:2309.09777}, 2023.

\bibitem[Wei et~al.(2024)Wei, Wang, Lu, Xu, Liu, Zhao, Chen, and Wang]{wei2024editable}
Yuxi Wei, Zi Wang, Yifan Lu, Chenxin Xu, Changxing Liu, Hao Zhao, Siheng Chen, and Yanfeng Wang.
\newblock Editable scene simulation for autonomous driving via collaborative llm-agents.
\newblock In \emph{CVPR}, 2024.

\bibitem[Wu et~al.(2023)Wu, Liu, Luo, Zhong, Chen, Xiao, Hou, Lou, Chen, Yang, et~al.]{wu2023mars}
Zirui Wu, Tianyu Liu, Liyi Luo, Zhide Zhong, Jianteng Chen, Hongmin Xiao, Chao Hou, Haozhe Lou, Yuantao Chen, Runyi Yang, et~al.
\newblock Mars: An instance-aware, modular and realistic simulator for autonomous driving.
\newblock In \emph{CAAI International Conference on Artificial Intelligence}, pages 3--15. Springer, 2023.

\bibitem[Xing et~al.(2025)Xing, Xia, Zhang, Chen, Yu, Liu, Liu, Wang, Shan, and Wong]{xing2025dynamicrafter}
Jinbo Xing, Menghan Xia, Yong Zhang, Haoxin Chen, Wangbo Yu, Hanyuan Liu, Gongye Liu, Xintao Wang, Ying Shan, and Tien-Tsin Wong.
\newblock Dynamicrafter: Animating open-domain images with video diffusion priors.
\newblock In \emph{ECCV}, pages 399--417. Springer, 2025.

\bibitem[Xing et~al.(2023)Xing, Wang, Zhou, Zhang, Yu, and Xu]{xing2023diffsketcher}
Ximing Xing, Chuang Wang, Haitao Zhou, Jing Zhang, Qian Yu, and Dong Xu.
\newblock Diffsketcher: Text guided vector sketch synthesis through latent diffusion models.
\newblock \emph{NeurIPS}, 36:\penalty0 15869--15889, 2023.

\bibitem[Yan et~al.(2024)Yan, Lin, Zhou, Wang, Sun, Zhan, Lang, Zhou, and Peng]{yan2024street}
Yunzhi Yan, Haotong Lin, Chenxu Zhou, Weijie Wang, Haiyang Sun, Kun Zhan, Xianpeng Lang, Xiaowei Zhou, and Sida Peng.
\newblock Street gaussians for modeling dynamic urban scenes.
\newblock \emph{arXiv preprint arXiv:2401.01339}, 2024.

\bibitem[Yang et~al.(2023)Yang, Ivanovic, Litany, Weng, Kim, Li, Che, Xu, Fidler, Pavone, et~al.]{yang2023emernerf}
Jiawei Yang, Boris Ivanovic, Or Litany, Xinshuo Weng, Seung~Wook Kim, Boyi Li, Tong Che, Danfei Xu, Sanja Fidler, Marco Pavone, et~al.
\newblock Emernerf: Emergent spatial-temporal scene decomposition via self-supervision.
\newblock \emph{arXiv preprint arXiv:2311.02077}, 2023.

\bibitem[Yang et~al.(2024)Yang, Wen, Ma, Mei, Li, Wei, Lei, Fu, Cai, Dou, et~al.]{yang2024drivearena}
Xuemeng Yang, Licheng Wen, Yukai Ma, Jianbiao Mei, Xin Li, Tiantian Wei, Wenjie Lei, Daocheng Fu, Pinlong Cai, Min Dou, et~al.
\newblock Drivearena: A closed-loop generative simulation platform for autonomous driving.
\newblock \emph{arXiv preprint arXiv:2408.00415}, 2024.

\bibitem[Yu et~al.(2024)Yu, Xing, Yuan, Hu, Li, Huang, Gao, Wong, Shan, and Tian]{yu2024viewcrafter}
Wangbo Yu, Jinbo Xing, Li Yuan, Wenbo Hu, Xiaoyu Li, Zhipeng Huang, Xiangjun Gao, Tien-Tsin Wong, Ying Shan, and Yonghong Tian.
\newblock Viewcrafter: Taming video diffusion models for high-fidelity novel view synthesis.
\newblock \emph{arXiv preprint arXiv:2409.02048}, 2024.

\bibitem[Zhang et~al.(2023)Zhang, Rao, and Agrawala]{zhang2023adding}
Lvmin Zhang, Anyi Rao, and Maneesh Agrawala.
\newblock Adding conditional control to text-to-image diffusion models.
\newblock In \emph{ICCV}, pages 3836--3847, 2023.

\bibitem[Zhang et~al.(2018)Zhang, Isola, Efros, Shechtman, and Wang]{zhang2018unreasonable}
Richard Zhang, Phillip Isola, Alexei~A Efros, Eli Shechtman, and Oliver Wang.
\newblock The unreasonable effectiveness of deep features as a perceptual metric.
\newblock In \emph{CVPR}, pages 586--595, 2018.

\bibitem[Zhang et~al.(2022)Zhang, Zhu, Zheng, Huang, Huang, Zhou, and Lu]{zhang2022beverse}
Yunpeng Zhang, Zheng Zhu, Wenzhao Zheng, Junjie Huang, Guan Huang, Jie Zhou, and Jiwen Lu.
\newblock Beverse: Unified perception and prediction in birds-eye-view for vision-centric autonomous driving.
\newblock \emph{arXiv preprint arXiv:2205.09743}, 2022.

\bibitem[Zhao et~al.(2024{\natexlab{a}})Zhao, Ni, Wang, Zhu, Huang, Chen, Wang, Zhang, Mei, and Wang]{zhao2024drivedreamer4d}
Guosheng Zhao, Chaojun Ni, Xiaofeng Wang, Zheng Zhu, Guan Huang, Xinze Chen, Boyuan Wang, Youyi Zhang, Wenjun Mei, and Xingang Wang.
\newblock Drivedreamer4d: World models are effective data machines for 4d driving scene representation.
\newblock \emph{arXiv preprint arXiv:2410.13571}, 2024{\natexlab{a}}.

\bibitem[Zhao et~al.(2024{\natexlab{b}})Zhao, Chen, Chen, Bao, Hao, Yuan, and Wong]{zhao2024uni}
Shihao Zhao, Dongdong Chen, Yen-Chun Chen, Jianmin Bao, Shaozhe Hao, Lu Yuan, and Kwan-Yee~K Wong.
\newblock Uni-controlnet: All-in-one control to text-to-image diffusion models.
\newblock \emph{NeurIPS}, 36, 2024{\natexlab{b}}.

\bibitem[Zheng et~al.(2025)Zheng, Chen, Huang, Zhang, Duan, and Lu]{zheng2025occworld}
Wenzhao Zheng, Weiliang Chen, Yuanhui Huang, Borui Zhang, Yueqi Duan, and Jiwen Lu.
\newblock Occworld: Learning a 3d occupancy world model for autonomous driving.
\newblock In \emph{ECCV}, pages 55--72. Springer, 2025.

\bibitem[Zhou et~al.(2024)Zhou, Simon, Peng, Mo, Zhu, Guo, and Zhou]{zhou2024simgen}
Yunsong Zhou, Michael Simon, Zhenghao Peng, Sicheng Mo, Hongzi Zhu, Minyi Guo, and Bolei Zhou.
\newblock Simgen: Simulator-conditioned driving scene generation.
\newblock \emph{arXiv preprint arXiv:2406.09386}, 2024.

\end{thebibliography}
\end{document}